# A Universal Logic Operator for Interpretable Deep Convolution Networks

**Kam Woh Ng\*, Lixin Fan† and Chee Seng Chan\***
\*Center of Image and Signal Processing, Faculty of Computer Science and Information Technology,
University of Malaya, Kuala Lumpur, Malaysia
†JD.COM American Technologies, Mountain View, CA
{kamwoh@siswa.um.edu.my;lixin.fan@jd.com;cs.chan@um.edu.my}

## Abstract

Explaining neural network computation in terms of probabilistic/fuzzy logical operations has attracted much attention due to its simplicity and high interpretability. Different choices of logical operators such as AND, OR and XOR give rise to another dimension for network optimization, and in this paper, we study the open problem of learning a universal logical operator without prescribing to any logical operations manually. Insightful observations along this exploration furnish deep convolution networks with a novel logical interpretation.

## Introduction

In this paper, we investigate a novel aspect of network configuration that has not been exploited before — we propose to learn the underlying *logic operations* between the neuron inputs and weights. Typical network computation, i.e. element-wise multiplications between the neuron inputs and weights, can be re-interpreted as an AND operation within the probabilistic logic frameworks. The alternative fuzzy XOR operation has also been studied and demonstrated advantages in its simplicity and high interpretability (Fan 2017). Naturally one may wonder whether different types of logical operators can be learned during the optimization of network *weights* and *bias terms*. This line of thinking sounds appealing, yet it also gives rise to some crucial questions that call for well thought out solutions e.g. i) how to parametrize different logical operations; ii) how to train a heterogeneous network with different types of logical operations; and iii) how to interpret the learned operations.

This paper provides insightful answers to these open questions by illustrating a universal logic operator (ULO) that is learned end-to-end with three image classification tasks. Empirically, it is shown that the proposed ULO not only learn to imitate prescribed logical operations, but at the same time it also learns complex behaviours that cannot be characterized by existing operators.

The rest of the paper is organized as follows. In the next section, we illustrate the proposed ULO following a brief review of different logical operators. Then, we demonstrate the effectiveness of ULO with image classification tasks and investigate in depth the characteristics of the learned logical operators. Finally, we conclude the paper with a discussion.

## Related Work

The vast majority of networks nowadays employ the element-wise multiplication between neuron inputs and weights. This computation can be re-interpreted as the logical AND operations as illustrated in this paper. We therefore do not discuss or compare any particular networks, instead, we merely use vanilla convolution/ResNet networks to study the prescribed logical AND operator. The alternative XOR operator was used in generalized hamming networks to quantify similarities between neuron inputs and weights (Fan 2017). In terms of probabilistic logical inferencing, all these networks are prescribed by a manually selected logical operation and the learning algorithms merely learn the *weights* and *bias terms*.

The most similar work is probably due to (Godfrey and Gashler 2017), which proposed to learn a generic parametrized activation function that characterizes a number of logical operations. However, the activation function was parametrized by a single parameter $\alpha$ which was unable to model all four typical logical operators considered in this paper (see Table 1). Furthermore, we re-implemented the $\alpha$-parametrized activation function and found that it was not comparable with the proposed ULO operation for image classification tasks reported in this paper[1].

It must be noted that there is abundant literature on probabilistic & fuzzy logic that are related to the probabilistic logical framework presented here. Due to the limited space, we refer readers to (Nilsson 1986; Belohlavek, Dauben, and Klir 2017) for thorough treatments of the topics.

## Neural Network with Probabilistic Logic Operators

Neural network computing has been inspired biologically and it is often interpreted from a signal processing point of view, where the neuron weights are considered related to non-linear mappings that transform input signals into semantically

---

[1](Godfrey and Gashler 2017) reported validation results for a network with the depth of logic layer fixed to two. All tests were carried out on five small datasets with no more than 5000 samples and 300 attributes.

| Inference rules $U(\varphi_x, \varphi_y)$ | Outputs $P(\varphi_c) = P(U(\varphi_x, \varphi_y))$ | Logical operator parameters |
|---|---|---|
| **ULO** $(\varphi_x, \varphi_y)$ | $\alpha xy + \beta y + \gamma x + b$ | $\alpha, \beta, \gamma, b$ **to be optimized** |
| AND $(\varphi_x, \varphi_y)$ | $xy$ | $\alpha = 1, \beta = 0, \gamma = 0, b = 0$ |
| OR $(\varphi_x, \varphi_y)$ | $x + y - xy$ | $\alpha = -1, \beta = 1, \gamma = 1, b = 0$ |
| XOR $(\varphi_x, \varphi_y)$ | $x + y - 2xy$ | $\alpha = -2, \beta = 1, \gamma = 1, b = 0$ |
| MP $(\varphi_x, \varphi_y)$ | $xy + (1-x)/2$ | $\alpha = 1, \beta = 0, \gamma = -0.5, b = 0.5$ |

Table 1: Comparison between our proposed universal logical operator (ULO) and four classical probabilistic logical inference rules (AND, OR, XOR, MP) under independence assumption. Note that $x = P(\varphi_x), y = P(\varphi_y)$ and MP stands for modus ponens, for which $P(\varphi_x) = P(A), P(\varphi_y) = P(B|A)$ and $P(\varphi_c) = P(B)$.

more pronounced features (Lecun, Bengio, and Hinton 2015; Goodfellow, Bengio, and Courville 2016).

In this paper, we re-interpret the element-wise neural computing as the probabilistic & fuzzy logical operations such as *AND, OR, XOR* and *modus ponens* (MP), and study the statistic characteristics of neuron outputs when different logical operators are applied. Moreover, we propose a *universal logical operation (ULO)* that learns to act in accordance with different logical inference rules. To our best knowledge, it is the first time for an end-to-end algorithm to learn logical operations instead of prescribing them manually for convolution neural networks.

## Inference with probabilistic logical operations

Within a probabilistic logical inference framework, input and output *strengths* of each neuron are treated as *probabilities* that the event $z$ in question is true i.e. $P(z = 1)$. For instance, for the output layers of a classification network, $P(z = 1)$ denotes the probability of detecting an object of interest. For the outputs of intermediate layers, the semantics of $P(z = 1)$ may represent the probability of finding certain patterns or object parts in the neuron inputs. For input layers, the statement $P(z = 1)$ may simply mean the probability that "the pixel value (intensity) is greater than 100". Without loss of generality, we use $P(\varphi)$ in this paper to denote the probability of a proposition $\varphi$ being true.

Often both the inputs and outputs of neurons do not lie in the interval $[0, 1]$, hence, are not valid probabilities. This technical nuisance can be trivially solved by normalizing neuron outputs with respect to a normalization constant or employing the logistic activation function to normalize $z$ as proper probabilities $P(\varphi) = \sigma(z) = \frac{1}{1+\exp(-z)}$ (see Chap 6 in (Goodfellow, Bengio, and Courville 2016)).

In this paper, we find it suffices to simply normalize $z$ at the output layer e.g. with a logistic function, while leaving un-normalized the network inputs, weights and outputs at all other layers, followed by applying the standard batch normalization (BN) to neuron outputs (Ioffe and Szegedy 2015). Given probabilities $P(\varphi_x), P(\varphi_y)$ of two propositions $\varphi_x, \varphi_y$, we are concerned with the probability $P(\varphi_c)$ of the composite proposition $\varphi_c = U(\varphi_x, \varphi_y)$, where $U()$ is a logical inference rule to be defined. Table 1 summarizes four different logic inference *rules* or *operators* typically used in probabilistic and fuzzy logic.

Among the four logical operators in Table 1, the probabilistic AND operation is the most commonly used in the majority of neural networks. Whereas probabilities of element-wise AND operation are computed as *multiplications* $x_k w_k$ with $k$ denotes the $kth$ (pixel) element of convolution (or fully connected) kernels, subsequently, element-wise probabilities are averaged and summed up with the bias term $B$ i.e. $z = \frac{1}{K}\mathbf{x} \cdot \mathbf{w} + B$. In this computations, element-wise probabilities are treated as *i.i.d* samples of the probability $P(\varphi_c)$ of $\varphi_c = \text{AND}(\varphi_x, \varphi_y)$. Finally, follow-up non-linear activation such as ReLU is applied to threshold the estimated probabilities of composite propositions i.e. $\max(0, z)$.

While majority of networks prescribe the AND operator, the XOR operator was also used in generalized hamming networks to quantify similarities between neuron inputs and weights (Fan 2017). On the other hand, probabilistic OR and modus ponens (MP) are not used in any neural networks. As illustrated in the next section, these operators characterize different statistical dependencies between the neuron inputs and outputs. It is this observation that inspires us to learn a universal logical operator without prescribing to any operators manually. In other words, we propose to *learn both neuron weights and the type of logical operators simultaneously*.

## Universal logical operator in neural networks

In this paper we propose a universal logical operator learning framework in which both the *types of logical operators* and neural network *weights & bias terms* are learned by an end-to-end algorithm. In order to parametrize different logical operations, we adopt the universal logical operator defined in Definition 1. The logic operation learned as such is therefore referred to as the *universal logical operator* (ULO) in the rest of the paper.

**Definition 1.** *A universal logical operator (ULO) is a parametric binary function $U : R \times R \to R$*

$$U_\theta(x, y) = \alpha xy + \beta y + \gamma x + b \qquad (1)$$

*in which $x, y \in R$ are un-normalized input probabilities and $\theta = \{\alpha, \beta, \gamma, b \in R\}$ is the set of parameters.*

Immediately one may notice that four of the prescribed logical operators in Table 1 are special forms of the universal logical operator, for instance, $\alpha = 1, \beta = 0, \gamma = -0.5$ and $b = 0.5$ for modus ponens.

Beside that, the learning of parameter $b$ is actually not required since it is absorbed in the bias term:

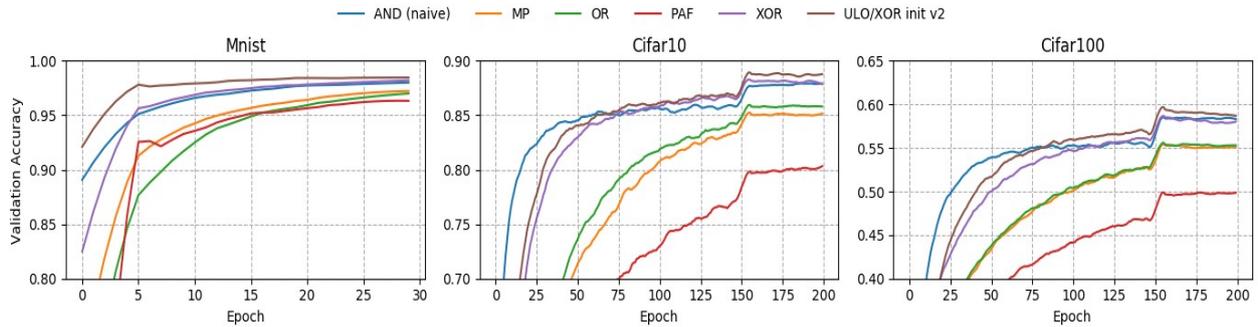

Figure 1: Image classification accuracies with different logical operators.

$z = \frac{1}{K}\sum_{k=1}^{K}(\alpha x_k w_k + \beta w_k + \gamma x_k + b) + B = \frac{1}{K}\sum_{k=1}^{K}(\alpha x_k w_k + \beta w_k + \gamma x_k) + (b+B)$. The remaining three elements then form a parameter vector $[\alpha, \beta, \gamma]^T$, of which AND, OR and MP are linearly independent[2]. It means any arbitrary ULOs can be decomposed as a linear combination of these prescribed operators i.e. a mixture model.

Equipped with the universal logical operator (ULO), the proposed learning algorithm merely has three additional logical parameters i.e. $\alpha, \beta, \gamma$ to optimize on top of the standard network weights ($w$) and bias terms ($B$)[3].

**Investigation of logical parameters**: the 3-tuple $(\alpha, \beta, \gamma)$ learned for each convolution & fully connected kernel can be normalized w.r.t. its length and plotted as a 3D point on an unit ball. As shown in Fig. 2, parameters of certain ULOs can be adjusted very close to e.g. AND operator although during the optimization there is no explicit incentive for these parameters to imitate prescribed operators. Equation (1) shows that these ULOs near AND/MP/OR/XOR operators behave in accordance to these logical inference rules. However, many ULOs do not exactly fit the prescribed logical operators and actually learn more complex behaviours that cannot be characterized by existing operators.

## Experimental results

We investigate the proposed *universal logical operator* (ULO) with three image classification tasks i.e. MNIST, CIFAR10 and CIFAR100. For comparison, the prescribed logical operators including AND, OR, XOR and modus ponens (MP) are used in different networks. Note that for AND operator, a vanilla 7-layered convolution and a 21-layered ResNet network are used for MNIST and CIFAR10/100, respectively. All other networks using different logical operators have the same architecture in terms of number of layers, filters, etc.

The fuzzy XOR operator proposed in (Fan 2017) is used to test XOR operator while OR and MP operators are implemented by ourself. We also re-implement the parametric activation function (PAF) proposed in (Godfrey and Gashler 2017) with the same network architecture used in this paper.

---

[2]XOR (=OR-AND) is linearly dependent.
[3]The source code of the learning algorithm will be made publicly available together with the publication.

Experimental results are divided into two aspects: i) overall performance in terms of accuracy; and ii) the analysis of learned logical operators in terms of distributions of logical parameters i.e. $(\alpha, \beta, \gamma)$, learned filters and estimated copulas. All experiments are repeated three times to obtain the average performances unless stated otherwise.

### Overall performances

Fig. 1 summarizes MNIST/CIFAR10/CIFAR100 classification accuracies for different logical operators. In general, it is observed that the performances of ULO, XOR and AND operators are similar with maginal yet consistent improvements observed for the ULO operator. On the other hand, the logical inferencing based on OR and modus ponens (MP) are less effective, with the final accuracies consistently lower than that of ULOs. This justifies that why they are not used in any neural networks. Correspondingly, the learning speeds also lag behind. Finally, the re-implemented parametric activation function (PAF) (Godfrey and Gashler 2017) is not comparable with the rest of the networks, especially for CIFAR10/100 classification in terms of accuracy and learning speed.

Note that for this comparison, we purposely did not fine-tune hyper-parameters (learning rate, optimization methods etc) for individual networks. Instead, a set of consistent hyper-parameters was used for all networks for a fair comparison, in which a learning rate decay from 0.1 to 0.01 scheduled at epoch 150 gave rise to performance jumps for all networks tested with CIFAR10/100.

### Convergence of logical parameters

In order to investigate the learned logical parameters i.e. $\alpha, \beta, \gamma$ (see Table 1), we plot *length-normalized* parameters as 3D points for convolution kernels at all layers over different learning epochs. Some example plots are summarized in Fig. 2, which shows that (2a) when randomly initialized, the scatter of logical parameters converge from the spherical surface to a circular one with parameter $\beta$ gradually reducing to a small value near 0.0; in case of parameters being initialized from AND parameters, $\beta$ is still reduced but the parameters cluster on two opposite arcs depending on the initialization.

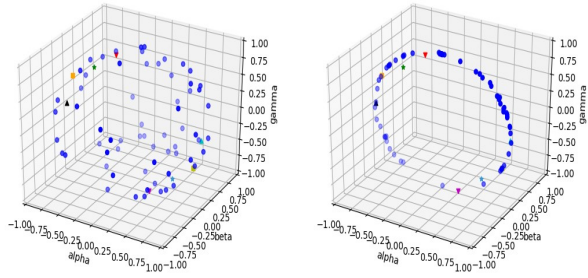

(a) Randomly initialized.

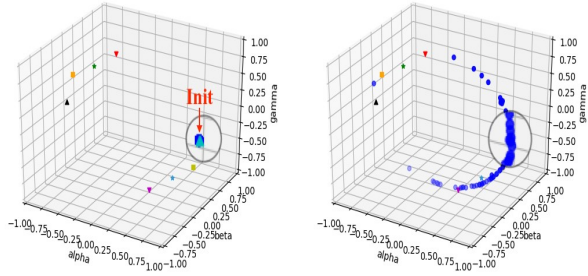

(b) Initialized as AND.

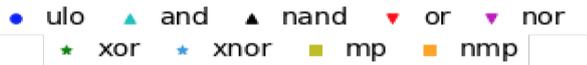

Figure 2: CIFAR10: Logical parameters of ULOs at last convolution layer. Left to right: epoch 0 and 200, respectively. Colour marks represent logical parameters for different operators (and/nand, or/nor, xor/xnor, mp/nmp are symmetric).

The convergence of parameters, as a result of simultaneous optimization of both neuron weights and logical parameters, is not enforced by the optimization objective function. Yet logical parameters automatically cluster near the prescribed logical operators such as AND. This observation demonstrated that (i) logical operators (or inference rules) are nothing but statistically stable patterns that can be learned from input data; (ii) although the learned ULOs appeared to favor the prescribed operators e.g. AND/XOR, the proposed framework admits flexibility in this choice by adapting logical operations to the input data; (iii) some logical parameters cannot be characterized by any prescribed operators as shown in Fig. 3. Nevertheless, these complex parameters can be decomposed as summations of different operators. Therefore, a mixed ULO operator is mathematically equivalent to a mixture of prescribed operators; and finally (iv) the learning of logical operators also established a strong connection between neural network optimization and logical inference rules learning, which is to be explored in future work.

## Discussion and conclusions

This paper explored a novel aspect of neural network computation by re-interpreting network inferencing in the lens of probabilistic/fuzzy logical operation. Concretely we proposed to parametrize different logical operations, which are learned simultaneously during the optimization of network weights and bias terms. The ULO learned as such is not confined to any prescribed probabilistic operators, although the majority of ULOs indeed learn to imitate or approximate the probabilistic AND operation. On the other hand, a small portion of dissident ULOs learn parameters far from any existing logical operators, and exhibit complex behaviour that can only be characterized by a mixture of prescribed operators.

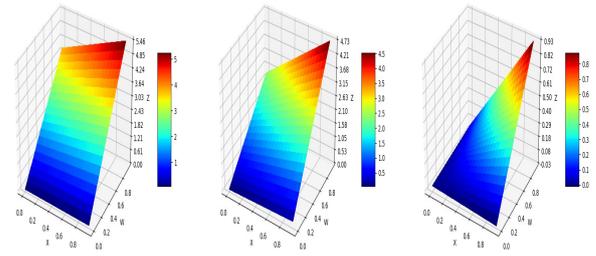

Figure 3: Function of mixed ULO. Left: ($\alpha = 2.40, \beta = 0.18, \gamma = 2.52$) = 7.26AND + 4.68MP - 0.18OR; Middle:($\alpha = 1.73, \beta = 0.02, \gamma = 4.08$) = 9.95AND + 8.2MP - 0.02OR; Right: Near AND operator ($\alpha = 1.03, \beta = -0.03, \gamma = 0.03$)

To our best knowledge, the work presented in this paper is the first demonstration whereas logical operation in network computation are learned by an optimization process. From an epistemic point of view, we empirically demonstrated that logical inference rules are nothing but statistically stable patterns that can be learned from input data. Moreover, the proposed framework admits flexibility in the choice of logical operations by adapting them to input data. We view this adaptation is the primary advantage and practical value of ULOs e.g. in hardware design. Finally it is our humble wish that the present paper will open a new direction which calls for follow up research to explore the connection between neural network optimization and logical operation learning.

## References

[Belohlavek, Dauben, and Klir 2017] Belohlavek, R.; Dauben, J.; and Klir, G. 2017. *Fuzzy Logic and Mathematics: A Historical Perspective*. Oxford University Press.

[Fan 2017] Fan, L. 2017. Revisit fuzzy neural network: Demystifying batch normalization and ReLU with generalized hamming network. In *NIPS*.

[Godfrey and Gashler 2017] Godfrey, L. B., and Gashler, M. S. 2017. A parameterized activation function for learning fuzzy logic operations in deep neural networks. *CoRR* abs/1708.08557.

[Goodfellow, Bengio, and Courville 2016] Goodfellow, I.; Bengio, Y.; and Courville, A. 2016. *Deep Learning*.

[Ioffe and Szegedy 2015] Ioffe, S., and Szegedy, C. 2015.

Batch normalization: Accelerating deep network training by reducing internal covariate shift. In *ICML*, 448–456.

[Lecun, Bengio, and Hinton 2015] Lecun, Y.; Bengio, Y.; and Hinton, G. 2015. Deep learning. *Nature* 521(7553):436–444.

[Nilsson 1986] Nilsson, N. J. 1986. Probabilistic logic. *Artificial Intelligence* 28(1):71 – 87.